\documentclass[10pt,twocolumn,letterpaper]{article}

\usepackage{iccv}              
\usepackage{multirow}
\usepackage{fix-cm}

\definecolor{iccvblue}{rgb}{0.21,0.49,0.74}
\usepackage[pagebackref,breaklinks,colorlinks,allcolors=iccvblue]{hyperref}
\usepackage[accsupp]{axessibility}  
\def\paperID{11900} 
\def\confName{ICCV}
\def\confYear{2025}

\title{Preacher: Paper-to-Video Agentic System}

\begin{document}

\author{Jingwei Liu$^{1,2}$\footnotemark[1] \quad Ling Yang$^{6}$\footnotemark[2]  \quad Hao Luo$^{2,3}$ \quad Fan Wang$^{2}$ \quad Hongyan Li$^{1,4, 5}$ \footnotemark[3] \quad Mengdi Wang$^{6}$ \footnotemark[3]\\
$^{1}$School of Intelligence Science and Technology, Peking University \\ $^{2}$DAMO Academy, Alibaba group, 310023, Hangzhou, China \\ $^{3}$Hupan Lab, 310023, Hangzhou, China\\ $^{4}$ National Key Laboratory of General Artificial Intelligence, Peking University \\$^{5}$ PKU-Wuhan Institude of Artificial Intelligence \\$^{6}$ Department of Electrical and Computer Engineering, Princeton University} 

%
\twocolumn[{
	\maketitle
	\setlength\tabcolsep{0.5pt}
	\centering
	\small
	\begin{tabular}{c}
        \includegraphics[width=0.99\textwidth]{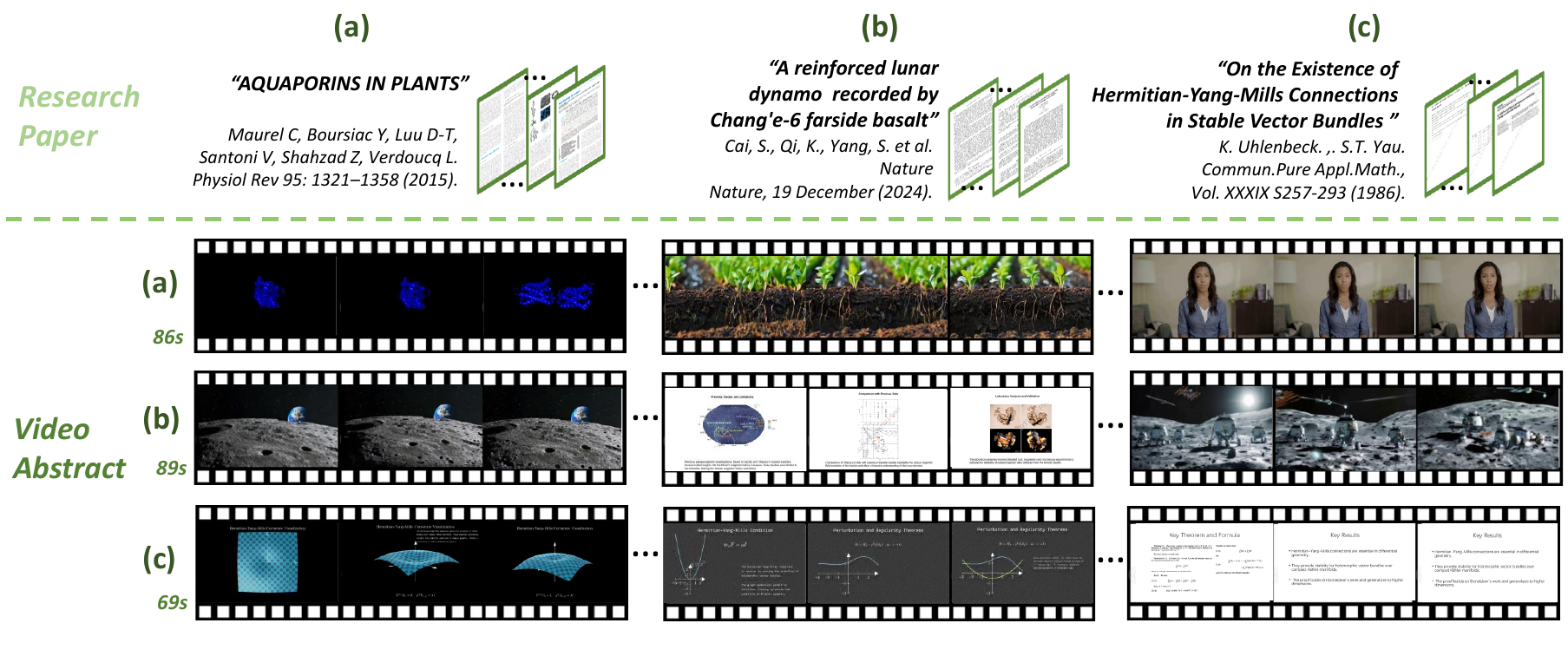}
	\end{tabular}
        \vspace{-0.2in}
	\captionof{figure}{Preacher can generate long video abstract conditioning on input paper with diverse topics.}
    \vspace{0.2in}
	\label{fig:teaser}
}]

\maketitle
\renewcommand{\thefootnote}{\fnsymbol{footnote}}
\footnotetext[1]{Work done during an internship at DAMO Academy.}
\footnotetext[2]{Contributed equally. Ling Yang, yangling0818@163.com}
\footnotetext[3]{Corresponding Authors: Hongyan Li, Mengdi Wang}
\begin{abstract}
The paper-to-video task converts a research paper into a structured video abstract, distilling key concepts, methods, and conclusions into an accessible, well-organized format. While state-of-the-art video generation models demonstrate potential, they are constrained by limited context windows, rigid video duration constraints, limited stylistic diversity, and an inability to represent domain-specific knowledge. To address these limitations, we introduce Preacher, the first paper-to-video agentic system. Preacher employs a top-down approach to decompose, summarize, and reformulate the paper, followed by bottom-up video generation, synthesizing diverse video segments into a coherent abstract. To align cross-modal representations, we define key scenes and introduce a Progressive Chain of Thought (P-CoT) for granular, iterative planning. Preacher successfully generates high-quality video abstracts across five research fields, demonstrating expertise beyond current video generation models. Code will be released at: \href{https://github.com/Gen-Verse/Paper2Video}{https://github.com/Gen-Verse/Paper2Video}
\end{abstract}    
\section{Introduction}
According to Scopus data \footnote{https://www.elsevier.com/products/scopus}, over three million scientific papers have been published since 2022, with an annual rise. As the volume of academic publications continues to grow, the need for effective dissemination and visibility has become increasingly critical. Among various dissemination strategies, video abstracts \citep{gupta2021integrated} offer a compelling means of communicating research findings by integrating visual and auditory elements, thereby enhancing comprehension and extending outreach. Studies have shown that papers accompanied by video abstracts receive 15\% more citations \citep{zong2019impact, ferreira2021audio, bonnevie2023video}. However, producing video abstracts remains resource-intensive, requiring both domain-specific expertise and professional video production skills, making it a costly process.

Given the recent advancements in generative artificial intelligence for video generation \citep{midjourney, pika, openai2024sora,tian2024videotetris}, developing an end-to-end video abstract generation model presents a compelling alternative to the high costs of manual production. While current methods can generate long-form videos exceeding 60 seconds \citep{openai2024sora}, they remain unsuitable for video abstract generation. First, contemporary methods exhibit inadequate capability in directly processing research papers containing embedded multimodal elements and long contexts.  Second, video generation frameworks trained on large-scale real-world video datasets \citep{schuhmann2021laion, schuhmann2022laion} exhibit rigid, homogeneous visual style, making them ill-suited for capturing the specialized representational demands of diverse academic disciplines.

To address these issues, we introduce Preacher, a novel paper-to-video agentic system integrating large multimodal models (LMMs), and specialized generative models. We introduce several key technologies in Preacher: \textbf{(i)} We construct a \textit{top-down and bottom-up} structure to support the complicated modality transition. In the top-down structure, Preacher decomposes and reformulates the paper as ``key scenes'', structured textual representations that encapsulate essential content while including visual descriptions to guide subsequent video generation. Serving as an intermediate bridge between textual and visual modalities, these key scenes ensure accurate content representation. In the bottom-up structure, key scenes are sequentially transformed into video segments, which are then assembled into a coherent video abstract. This structure enables precise collaboration between LMMs and generative models, effectively mitigating context window limitations while ensuring high-quality video generation. \textbf{(ii)} To enhance key scene planning and counteract the performance degradation of LMMs when handling long contexts \citep{li2024loogle, li2024long} or low-level detailed planning \citep{yang2024mastering,tian2024videotetris}, we introduce a progressive chain of thought (P-CoT). This method enables incremental fine-grained planning, improving coherence and scene accuracy. \textbf{(iii)} Preacher integrates video generation models with distinct styles, alongside Python-based professional visualization tools, allowing for the adaptive presentation of specialized content in the most appropriate video format. By aligning content planning with style selection, Preacher ensures that domain-specific concepts are effectively conveyed in academically relevant visual representations.

Through the top-down and bottom-up structure, multi-agent collaboration is effectively facilitated, generating high-quality video abstracts. To conduct a systematic evaluation, we employ an LMM to comprehensively evaluate the generated video abstracts across multiple dimensions, including accuracy, professionalism, aesthetic quality, and alignment with the input paper. We separately evaluate key scene planning and video generation quality, enabling direct comparison with alternative approaches. Preacher was tested on papers from five research fields and compared against state-of-the-art LMMs and video generation frameworks. Empirical results indicate that Preacher outperforms existing methods in both planning and generation, further substantiating its efficacy and applicability.

Our main contributions are as follows:
\begin{itemize}
    \item  We introduce Preacher, the first agentic system  to autonomously convert papers into video abstracts.
    \item  We develop a top-down and bottom-up structure to augment agent collaboration, and introduced key scenes bridging the gap between disparate modalities, with P-CoT enabling fine-grained key scene planning.
    \item  We validate Preacher across five research fields, demonstrating its capability as an end-to-end solution that mitigates the high costs of manual video production and enhances knowledge dissemination.
\end{itemize}

\section{Related Work}
\subsection{Automatic Knowledge Summary}
With the advancements in LMMs \citep{wang2024qwen2,yang2025mmada}, including enhanced text comprehension and expanding context lengths \citep{VeryLong, dubey2024llama, LongRoPE}, research has focused on leveraging LMMs for automated knowledge extraction and summarization \citep{scherbakov2024emergence, wu2025survey, letiche5110658literature}. \cite{jin2024comprehensive} propose an end-to-end review-generation pipeline with preprocessing, modeling, and evaluation stages. Similarly, AutoSurvey \citep{wangautosurvey} utilizes LMMs to retrieve and synthesize existing literature, while \citet{tian2024overview} introduced techniques such as clustering, dimensionality reduction, and stepwise prompting to enhance knowledge extraction from research papers. Agentic systems have also been explored for automated paper reviewing \citep{zhou2024llm, jin2024agentreview}.
However, existing methods primarily output textual summaries, which often fail to effectively convey key visual elements such as figures, charts, and experimental workflows, limiting the accessibility and impact of research findings. To address this limitation, we propose automatically generated video abstracts as a more intuitive and comprehensive alternative to traditional textual summaries.
\subsection{Conditional Video Generation}
Conditional video generation has been a core topic in machine learning research. Early models were constrained to 16-frame outputs \citep{ho2022video}, with subsequent approaches incorporating text-to-image diffusion models \citep{rombach2022high,yang2024mastering,yang2023improving,wang2025rectified} to extend generation length \citep{khachatryan2023text2video, weng2024art}. Beyond text-based conditioning, image-conditioned generation has emerged as a complementary approach. VideoComposer \citep{wang2023videocomposer} integrates images as control signals into the diffusion process, and VideoCrafter2 \citep{chen2024videocrafter2} leverages CLIP-derived textual and visual embeddings for cross-attention. However, these methods primarily produce simple motions and struggle with frame consistency in extended sequences, which are further improved in StreamingT2V \citep{henschel2024streamingt2v} and VideoTetris \citep{tian2024videotetris}. Recent efforts have addressed these limitations by adopting regression-based conditioning, leveraging previous frames for improved temporal coherence in long-form video synthesis \citep{zhang2023i2vgen, zeng2024make, xing2025dynamicrafter,tian2024videotetris}.

While closed-source models remain state-of-the-art in performance \citep{openai2024sora, pika, luma}, enabling generation at scales of tens of seconds, they cannot process research papers as direct inputs and fail to accommodate the stylistic diversity required for video abstracts. To bridge this gap, we integrate LMMs with a suite of heterogeneous video generation tools, forming a collaborative framework capable of processing research papers as input and producing long-form video abstracts in varied, contextually appropriate styles.
\vspace{-0.5em}
\subsection{Agentic Systems}
Recent advancements in LMM-based agentic system have demonstrated reasoning and planning capabilities approaching human-level performance, aligning with the expectations for autonomous agents—systems capable of perceiving environments, making decisions, and executing actions. Compared to single-agent approaches \citep{achiam2023gpt,touvron2023llama}, agentic systems harness collective intelligence and specialized expertise, enabling them to address complex challenges, including advanced programming tasks \citep{li2023camel, dong2024self, hongmetagpt} and planning in physical environments \citep{dasgupta2022collaborating, song2023llm, huang2023inner, guo2024large}. Several studies explore agentic systems to enhance the capabilities of generative models \citep{allioui2022multi, gal2024comfygen, wang2025genartist}. In video generation, DreamFactory \citep{xie2024dreamfactory} employs multi-agent collaboration and keyframe iteration to ensure consistency and style in long-form videos, while Mora \citep{yuan2024mora} integrates human-in-the-loop feedback to refine output quality. SPAgent \citep{tu2024spagent} autonomously orchestrates tools for video generation and editing through a structured three-step framework. Unlike existing approaches, our methodology advances agentic systems by introducing enhanced collaborative mechanisms, enabling the execution of cross-modal tasks that exceed the capabilities of a single agent.
\vspace{-0.5em}
\section{Preliminary}
Let $P$ represent a complete and standardized academic paper, consisting of text, equations, figures, and tables. A video $V$is represented as a sequence of frames:$V={F_1, F_2,...,F_T}$, where each $F_t$ corresponds to an image at time step $t$. Video abstracts may even incorporate multiple styles as a special kind of video \citep{gupta2021integrated} \footnote{https://www.animateyour.science/post/8-ways-to-make-a-video-abstract}. For clarity, we define $V$ specifically as a video abstract:$V = {V^{s_1}_1,..., V^{s_n}_i, ..., V^{s_N}_H}$, where $s_n \in \mathbb{S}$ and $\mathbb{S}$ is the space of all possible video abstract styles, and $V^{s_n}_i$ represents a segment of a video abstract with a specific style. 

Formally, we aim to learn a generative model $G$ that maps $P$ into video abstract $V$ within the video space $V = G(P)$. We construct an agentic system, decomposing  $G$ into a set of agents ${\mathcal{A}}$, each dedicated to a distinct subtask. These agents collaborate with each other, ensuring the generation of stylistically diverse video abstracts.
\vspace{-0.5em}
\section{Preacher}

\begin{figure*}[ht]
\centering
\includegraphics[width=0.95\linewidth]{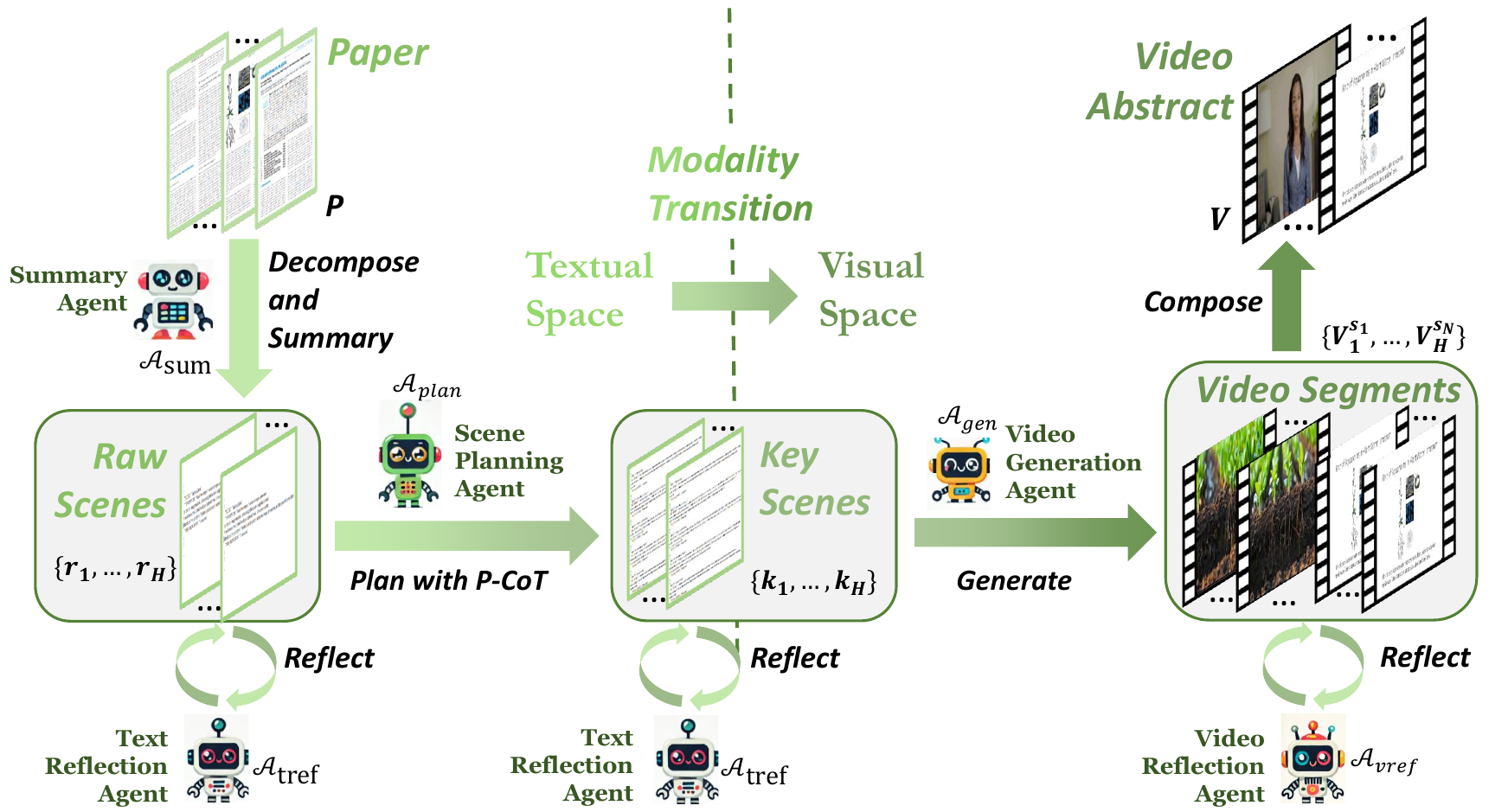}
\vspace{-0.07in}
\caption{\textbf{Overview of Preacher}. The summary agent $\mathcal{A}_{\text{sum}}$ decomposes and summarizes the paper into $H$ raw scenes. Subsequently, the planning agent $\mathcal{A}_{\text{plan}}$ and video generation agents $\mathcal{A}_{\text{gen}}$ then iteratively process these scenes, generating $H$ corresponding video segments, which are subsequently assembled into a complete video abstract. For clarity, $\mathcal{A}_{\text{form}}$ is omitted, with the detailed workflow provided in Appendix B.3.}
\vspace{-1em}
\label{fig:mainframe}
\end{figure*}
Preacher is a paper-to-video agentic system integrating LMMs, LMMs, and diverse generative models. \cref{sec_3_1} outlines the architecture of the system and the specialization of the agents. \cref{sec_3_2} details the key scene planning and presents the progressive chain of thought to improve planning accuracy. Finally, \cref{sec_3_3} introduces how Preacher utilizes key scenes to generate video abstracts.
\vspace{-0.5em}
\subsection{Overview of Preacher}
\label{sec_3_1}
\vspace{-0.2em}
\paragraph{Top-Down and Bottom-Up Structure}
Most existing cross-modal agentic systems employ a unified multi-step pipeline for cross-modal tasks \citep{yuan2024mora, zhu2024intelligent, tu2024spagent, zheng2025pptagent}. However, their performance is heavily dependent on existing text-to-visual generation models, making them ineffective for processing highly complex inputs. Inspired by prior research \citep{qian2023communicative, hongmetagpt, dong2024self}, we decompose and summarize input papers before feeding them into generative models. While these summaries improve compatibility, the resulting videos are often low-quality and semantically hollow. This limitation arises from insufficient detail in the summaries, preventing accurate reconstruction of key content, and from granularity constraints that hinder contemporary models' ability to fully leverage CLIP-based cross-modal mechanisms \citep{radford2021learning}.

To address these challenges, we introduce a top-down and bottom-up framework, inspired by the U-Net architecture \citep{ronneberger2015u}. In the top-down phase, the paper undergoes decomposition and summarization into multiple raw scenes, each encapsulating core content while omitting fine details, serving as anchors for content segmentation. Analogous to the U-Net encoder, where spatial resolution is reduced while feature depth increases, we perform structured planning on these downsampled raw scenes, enriching them with higher-dimensional information. 

Unlike prompt augmentation \citep{bodur2024prompt}, Preacher’s planning process continuously references the original paper, ensuring that the generated content maintains precise semantic alignment with the source material. The planning results termed key scenes, not only enhance compatibility with Preacher's generation tools but also embed rich semantic information, providing multi-dimensional guidance for subsequent generative processes \cref{sec_3_2}.

Within the bottom-up structure, agents equipped with video generation tools reconstruct the key scenes, generating both video and corresponding audio. Each video segment is synthesized from these elements, and upon completion, all segments are integrated into the final video abstract. 

\vspace{-1em}
\paragraph{Agent Specialization}
Agent specialization allows agents to collaborate on tasks that a single agent cannot complete. We have six agents in the Preacher: the Summary Agent $\mathcal{A}_{\text{sum}}$, the Format Agent $\mathcal{A}_{\text{form}}$, the Scene Planning Agent $\mathcal{A}_{\text{plan}}$, the Text Reflection Agent $\mathcal{A}_{\text{tref}}$, the Video Reflection Agent $\mathcal{A}_{\text{vref}}$, the Video Generation Agent $\mathcal{A}_{\text{gen}}$. A brief description of agents follows, with more details in \cref{sec4_id}.

\begin{itemize}
    \item Summary Agent $\mathcal{A}_{\text{sum}}$:  This agent employs LMMs, such as GPT-4o \citep{achiam2023gpt} and Gemini \citep{team2023gemini}, to understand, decompose and summary the input paper. 
    \item Format Agent $\mathcal{A}_{\text{form}}$: This agent employs LLMs, such as Llama \citep{touvron2023llama, touvron2023llama2} to format the output from $\mathcal{A}_{\text{sum}}$, ensuring the output of $\mathcal{A}_{\text{sum}}$ is correctly structured as raw scenes.
    \item Scene Planning Agent $\mathcal{A}_{\text{plan}}$: This agent employs LMMs, same as $\mathcal{A}_{\text{sum}}$, and its task is to provide a more detailed plan for each raw scene. 
    \item Rule-based Reflection Agents $\mathcal{A}_{\text{tref}}$ and $\mathcal{A}_{\text{vref}}$: There are two reflection agents in Preacher: $\mathcal{A}_{\text{tref}}$ and $\mathcal{A}_{\text{vref}}$. They are both based on LMMs. 
    \item Video Generation Agent $\mathcal{A}_{\text{gen}}$: $\mathcal{A}_{\text{gen}}$ is composed of LMMs and video generation tools, designed to generate videos with key scenes. $\mathcal{A}_{\text{gen}}$ is equipped with variable video generation tools: the Python package, text-to-image models \citep{rombach2022high, tywx, ramesh2022hierarchical}, text-to-video models \citep{rombach2022high, tywx, pika, luma, openai2024sora}, talking heads generation models\citep{tavus, gao2023high,cui2024hallo2}.
\end{itemize} 

\begin{figure*}[h]
\centering
\includegraphics[width=0.99\linewidth]{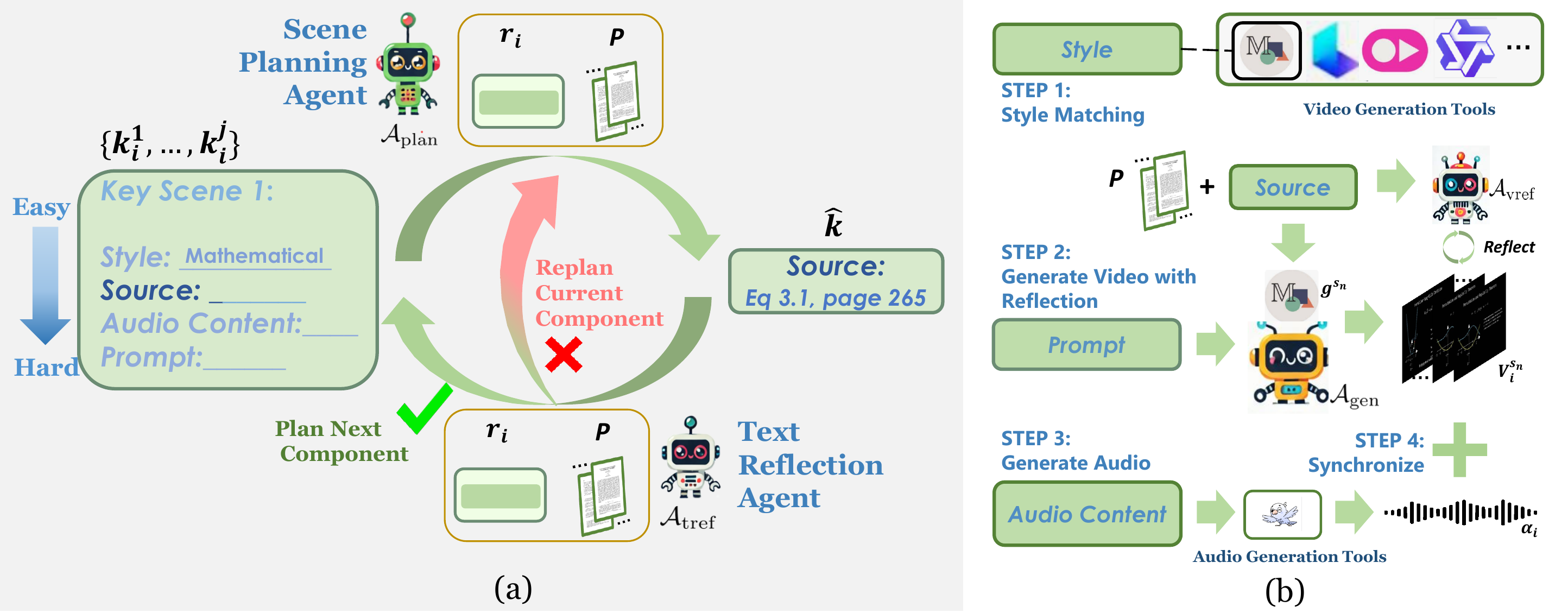}
\vspace{-0.1in}
\caption{(a) A schematic representation of the progressive chain of thought. The key scenes consist of multiple components requiring systematic planning. The Scene Planning Agent $\mathcal{A}_{\text{plan}}$ devises a structured plan for each component, which is then evaluated by the Text Reflection Agent 
$\mathcal{A}_{\text{tref}}$. Based on the reflection outcome, $\mathcal{A}_{\text{plan}}$ either advances to the next component using the existing plan or revises the current component, iterating this process until all components are effectively planned. (b) The Generation Agent $\mathcal{A}_{\text{gen}}$ utilizes the key scenes to synthesize video segments. The elements enclosed within the green frame represent the structured components of the key scenes.}
\vspace{-1em}
\label{fig:trick}
\end{figure*}
\subsection{Automatic Planning of Key Scenes}

\label{sec_3_2}
\paragraph{Progressive CoT Planning}
As illustrated in \cref{fig:trick}, key scenes comprise essential elements, including duration, video style, audio content, video prompts, and corresponding sources (e.g., specific sections, figures, or equations from the original paper). Serving as an intermediary between the top-down and bottom-up structures, key scenes facilitate seamless cross-modal representation alignment. To ensure effective planning, we employ a multi-agent collaboration framework to systematically refine key scenes.
\vspace{-0.7em}
\begin{align}
    \label{E_1}
    \{r_1, r_2, ... ,r_H\} \leftarrow \mathcal{A}_{\text{form}}(\mathcal{A}_{\text{sum}}(P)) \\
    k_i \leftarrow \mathcal{A}_{\text{plan}}(r_i, P), i = 1, 2, ... , H
\end{align}

The quality of a video abstract is highly contingent on the effective planning of key scenes. However, LMMs exhibit degraded performance in low-level planning tasks, particularly when handling long-context dependencies \citep{li2024long}. A common issue is the generation of partially inappropriate scenes, which, despite evaluation and re-execution, may correct prior errors while inadvertently introducing new ones. Additionally, after multiple rounds of re-planning, LMMs may deviate from the original task objective due to accumulated contextual drift from repeated reflections.

To address these limitations, we introduce the Progressive Chain of Thought (P-CoT), a specialized CoT framework that incorporates reflection mechanisms to enhance planning coherence. As illustrated in \cref{fig:trick}(a), when planning across  $J$ components, tasks are assigned to $\mathcal{A}_{\text{plan}}$ sequentially in a simple-to-complex order, with one component planned at a time.
\vspace{-0.3em}
\begin{equation}
\small
\label{E_3}
    \hat{k} \leftarrow 
    \begin{cases} 
        \mathcal{A}_{\text{plan}}(r_i, P) & \text{if } j = 1, \\
        \mathcal{A}_{\text{plan}}(r_i, P, \{ k_i^n | 1 \leq n \leq j-1 \}) & \text{else}.
    \end{cases}
\end{equation}
where $\{ k_i^n : 1 \leq n \leq j-1 \}$ are the approved components in the $i_{th}$ key scene and $\hat{k}$ is the plan for the current component. The agent focuses on the $\hat{k}j_i$ until it has been approved by $\mathcal{A}_{\text{tref}}$. Once $\mathcal{A}_{\text{tref}} (\hat{k})$ is approved, it is fixed and passed to $\mathcal{A}_{\text{plan}}$ to plan the subsequent components:
\vspace{-0.5em}
\begin{equation}
\label{E_4}
\small
        k^j_i \leftarrow \hat{k}, j \leftarrow j + 1 \quad \text{if } \mathcal{A}_{\text{tref}} (\hat{k}) \rightarrow \text{Approved}
\end{equation}
\vspace{-0.2em}
If disapproved, $\mathcal{A}_{\text{tref}}$ provides reflection to $\mathcal{A}_{\text{plan}}$ for re-planning \cref{E_3}. This iterative process continues until all components within the key scene are approved. The progressive complexity approach mitigates the challenges of intricate scene planning while addressing inconsistencies arising from iterative plannings.

\vspace{-1.2em}
\paragraph{Structured Communication between Agents}
While natural language communication between agents offers convenience, it is inherently unstable, as LMMs may introduce ambiguities or incomplete responses \citep{hongmetagpt}. In Preacher, incompleteness in natural language-driven planning can substantially impair the effectiveness of the subsequent video generation agent. $\mathcal{A}_{\text{gen}}$. 

To address this issue, we implement a structured fill-in task format, where the Format Agent $\mathcal{A}_{\text{form}}$ populates predefined dictionaries with the appropriate content. As illustrated in \cref{fig:mainframe}, both raw scenes and key scenes are stored as structured \textit{json} files, ensuring consistency and reliability. Additionally, human users retain the flexibility to manually create or modify \textit{json} files, either substituting the top-down structure or refining existing scene plans as needed.
\begin{figure*}[t]
\centering
\includegraphics[width=0.99\linewidth]{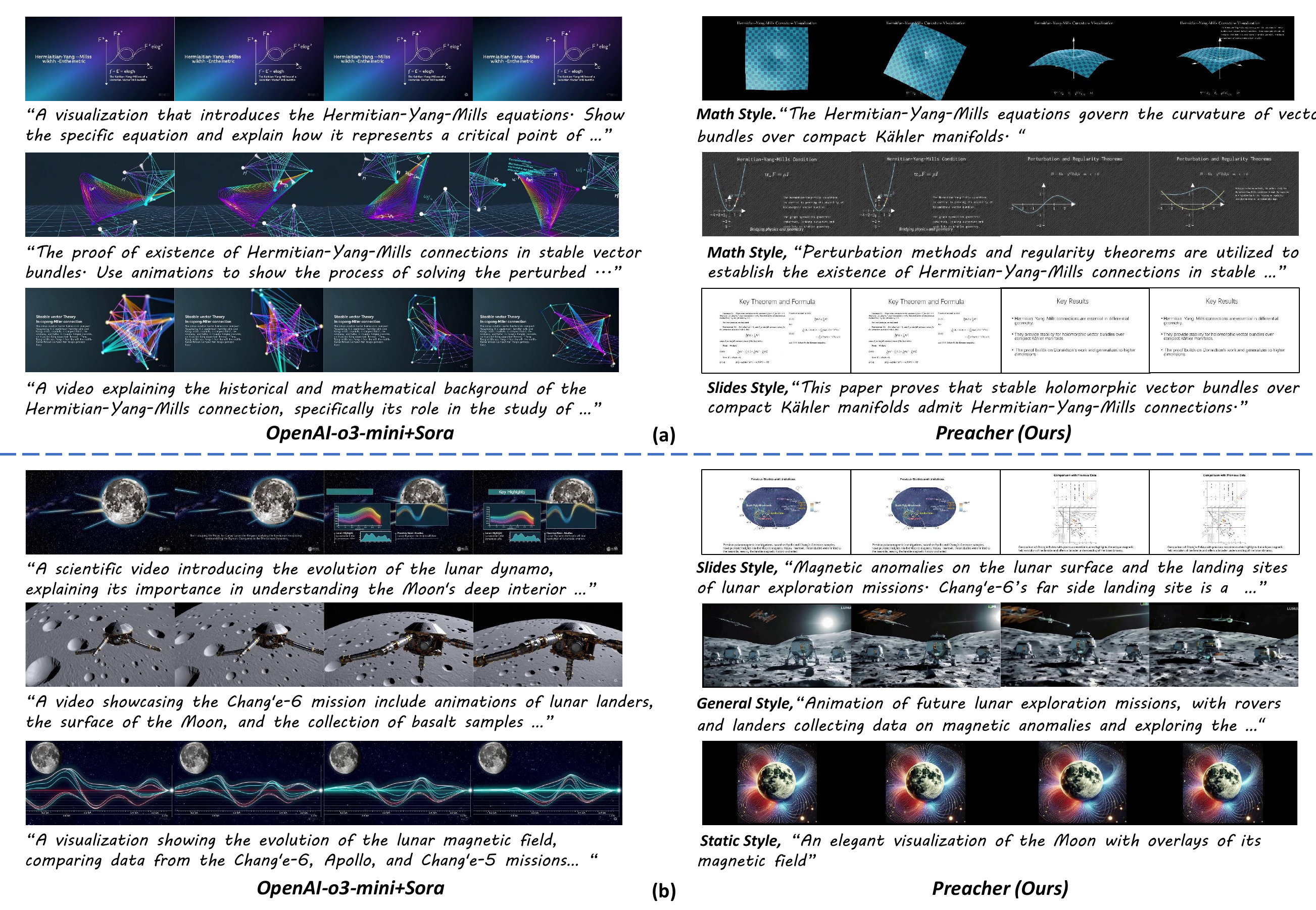}
\vspace{-0.07in}
\caption{Comparison of the output videos generated by OpenAI o3-mini \citep{jaech2024openai} + Sora \citep{openai2024sora} and Preacher. The upper and lower sections present frames in video abstract generated from (a) ``On the Existence of Hermitian-Yang-Mills Connections in Stable Vector Bundles'' \citep{yms}  and (b) \textit{``A reinforced lunar dynamo recorded by Chang'e-6 farside basalt''\citep{cai2024reinforced}}, respectively. A selection of frames has been chosen for demonstration.}
\label{fig:showcase}
\end{figure*}
\vspace{-0.25em}
\subsection{Generating Professional Video Abstracts}
\label{sec_3_3}
While existing video generation models \citep{openai2024sora} are proficient in generating conventional scenes and motions, they encounter challenges in producing content that requires specialized knowledge, such as mathematical concepts or the structural representation of specific molecules. To mitigate this challenge, we have integrated multiple video generation tools in $\mathcal{A}_{\text{gen}}$. Upon acquiring key scene with style $s_n$, $\mathcal{A}_{\text{gen}}$ initially selects the appropriate video generation tool $g^{s_n}$. Six video styles are supported in Preacher: ``talking heads,'' ``general,'' ``static concept,'' ``molecular visualization,'' ``slides,'' and ``mathematics.'' More details about these video styles can be found in Appendix B.1. If the style of the video in the key scene is ``molecular visualization'', ``slides'' or ``mathematics'', we utilize LMMs to generate the corresponding Python code and execute it. Given the inherent susceptibility of this process to execution failures, $\mathcal{A}_{\text{tref}}$ iteratively reviews and refines the generated code to enhance its executability, ensuring successful script execution:
\vspace{-0.4em} 
\begin{equation}
\label{E_5}
\small
        V^{s_n}_i=g^{s_n}(\tau), \tau, s_n \in k_i
\end{equation}
where $\tau$ denotes the code or the prompt, and $g^{s_n}$ represents the video generation model or the execution of Python code. 

To enhance the quality of the video, $\mathcal{A}_{\text{vref}}$ conducts a thorough evaluation of the generated video segment. The evaluation criteria include: (i) Accuracy, (ii) Professionalism, (iii) Alignment between the video content and the paper. If the video segment does not meet the required standards, $\mathcal{A}_{\text{vref}}$ will directly modify $\tau$  and initiate the regeneration process \cref{E_5}.

Once the video segment is generated, $\mathcal{A}_{\text{gen}}$ will generate the corresponding audio ${\alpha}_i$ and integrate it with the video segment. This process is repeated $H$ times and video segments are concatenated to form the final video:

\begin{equation}
\label{E_6}
\small
        V \leftarrow \bigoplus_{i=1}^{H} \tilde{V}^{s_n}_i, \tilde{V}^{s_n}_i \leftarrow \text{syn}({V}^{s_n}_i, {\alpha}_i)
\end{equation}
where $\text{syn}(\cdot)$ and $\bigoplus$ represents the synchronization and video composition, respectively.

\vspace{-0.5em}
\section{Experiments}
\begin{table*}[!ht]
    \small
    \centering
    \caption{Performance comparisons on forty videos in terms of ten metrics. We report mean values and standard error. The best is in bold, while the second best is underlined.}
    \resizebox{0.99\textwidth}{!}{
    \label{tab:overall}
    \begin{tabular}{l|cccc|cccc|c|c}
    \toprule
    \multirow{2}{*}{\textsc{\textbf{Method}}} & 
    \multicolumn{4}{c|}{\textsc{\textbf{GPT Evaluation}}} &
    \multicolumn{4}{c|}{\textsc{\textbf{Human Evaluation}}} &
    \multicolumn{1}{c|}{\multirow{2}{*}{\textsc{\textbf{CLIP}} $\uparrow$ }}&
    \multicolumn{1}{c}{\multirow{2}{*}{\textsc{\textbf{AE}} $\uparrow$ }}
    \\
    & \textbf{Accuracy} $\uparrow$ & \textbf{Professionalism} $\uparrow$  & \textbf{Aesthetic} $\uparrow$  & \textbf{Alignment} $\uparrow$ & \textbf{Accuracy} $\uparrow$ & \textbf{Professionalism} $\uparrow$ & \textbf{Aesthetic} $\uparrow$ & \textbf{Alignment} $\uparrow$ & \\
    \midrule
    OpenAI-o3-mini \citep{O3} + StreamingT2V \citep{henschel2024streamingt2v} & 3.35(0.98) & 4.03(0.87) & 4.00(0.77) & 3.60(0.91) & 3.13(0.93) & 3.83(0.96) & 3.10(1.21) & 3.87(0.86) & 0.23(0.04)  & 4.99(0.67)\\
    OpenAI-o3-mini  + Wan 2.1-14B \citep{tywx} & 3.75(0.43) & \underline{4.53}(0.48) & 4.15(0.41) & \underline{4.33}(0.69) & 3.63(0.91) & 4.45(0.49) & \textbf{4.33}(0.51) & 4.23(0.69) & \underline{0.29}(0.07)  & \underline{5.29}(0.47)\\
    OpenAI-o3-mini  + Kling 1.6 \citep{kl} & 3.70(0.61) & 4.18(0.79) & 3.98(0.73) & 4.05(0.83) & 3.40(1.06) & 4.23(0.87) & 4.13(0.69) & 3.78(0.89) & 0.26(0.07)  & 5.18 (0.63)\\
    OpenAI-o3-mini + Sora\citep{openai2024sora}  & \underline{4.33}(0.94) & 4.45 (0.49) & \textbf{4.18}(0.67) & 4.30(0.73) & \underline{3.88}(0.86) & 4.50(0.67) & \underline{4.30}(0.49) & \underline{4.38}(0.59) & \textbf{0.31}(0.06)  & \textbf{5.31}(0.53)\\
    \midrule
    Preacher (\textbf{Ours}) & \textbf{4.50}(0.55) & \textbf{4.63}(0.44) & \underline{4.17}(0.69) & \textbf{4.35}(0.98) & \textbf{4.80}(0.46) & \textbf{4.78}(0.46) & 4.25(0.58) & \textbf{4.75}(0.43) & 0.26(0.09)  & 5.20(0.83)\\
    \bottomrule
    \end{tabular}
}
\end{table*}
\subsection{Experimental Setup}
\label{sec5_1}
\textbf{Benchmark.} To assess the effectiveness of Preacher, we constructed a benchmark dataset comprising 40 research papers spanning five distinct fields: Mathematics, Molecular Biology, Geology, Machine Learning, and Climate Science. These papers were randomly selected using GPT-4o \citep{zheng2023chatgpt}, and the complete list is provided in Appendix A.

As no directly comparable baseline exists, we establish an end-to-end paper-to-video generation pipeline by integrating an LMM with a video generation model. Specifically, OpenAI-o3-mini-high \citep{O3} serves as the scene decomposition module, segmenting the input paper into multiple key scenes, while state-of-the-art video generation models synthesize 5-second video segments from these scenes. We evaluate multiple video generation models, including the open-source methods StreamingT2V \citep{henschel2024streamingt2v}, VideoTetris \citep{tian2024videotetris} and Wan-2.1-t2v-14B \citep{tywx}, as well as the closed-source models OpenAI Sora \citep{openai2024sora} and Kling 1.6 \citep{kl}. To evaluate the Preacher's ability to plan key scenes, we also employed other LMMs to directly plan key scenes and use GPT-4o as the judge.

\textbf{Evaluation Metrics.}
We utilize GPT-4 to evaluate the quality of the final video, with GPT-4 providing scores ranging from 1 to 5 in the following aspects: (i) Accuracy: Correctness of the video content, free from errors. (ii) Professionalism: Use of domain-specific knowledge and expertise. (iii) Aesthetic Quality: Visual appeal, design, and overall presentation. (iv) Alignment with the Paper: Semantic Alignment with the paper. Additionally, we use the CLIP text-image similarity score (CLIP) \citep{radford2021learning}  and Aesthetic Score (AE) \citep{schuhmann2022laion} to evaluate the consistency with the prompt and aesthetic quality. For key scene evaluation, we introduce similar metrics: Accuracy, Professionalism, Compatibility, and Alignment. Here, Compatibility measures the feasibility of directly generating scenes, reflecting the effectiveness of the planning process.
All metrics are computed individually, and the results are averaged across all videos for overall evaluation. For quantitative analysis, we sample 60 frames per video to ensure consistency across evaluations.

\textbf{Implementation Details.}
\label{sec4_id}
Preacher primarily integrates existing APIs and Python scripting, with no GPU requirement. We use Gemini-2.0-flash \citep{team2023gemini} as the LMM in $\mathcal{A}_{\text{sum}}$ and $\mathcal{A}_{\text{plan}}$, as Gemini's API allows the direct upload of an entire encoded PDF as context. GPT-4o is utilized for $\mathcal{A}_{\text{form}}, \mathcal{A}_{\text{tref}},$ and $\mathcal{A}_{\text{vref}}$, where the PDF is processed through an assistant pipeline \footnote{https://platform.openai.com/docs/assistants/overview}. For $\mathcal{A}_{\text{gen}}$, we employ specialized Python libraries to generate professionally styled videos, specifically using: manim for mathematical animations, python-pptx for slide-based visualizations, and Pymol for molecular visualization. Furthermore, we employ Wan-2.1-t2i-turbo \citep{tywx} as the text-to-image approach, CosyVoice2 \citep{du2024cosyvoice} as the text-to-speech approach, Luma \citep{luma} as the text-to-video approach, and Tavus \citep{tavus} as the talking-head generation approach. Specific details regarding video styles and implementation methods can be found in Appendix B.1. 

\vspace{-0.8em}
\begin{table}[!ht]
    \small
    \centering
    \caption{Performance comparisons on key scenes on four metrics. We report mean values and standard error. The best is in bold, while the second best is underlined.}
    \vspace{-0.5em}
    \resizebox{0.49\textwidth}{!}{
    \label{tab:overall2}
    \begin{tabular}{l|cccc}
    \toprule
    \textsc{\textbf{Method}} & 
     \textbf{Accuracy} $\uparrow$ & \textbf{Professionalism} $\uparrow$  & \textbf{Compatibility} $\uparrow$  & \textbf{Alignment} $\uparrow$ \\
    \midrule
    GPT-4o \citep{zheng2023chatgpt} & 4.05(0.81) & 4.30(0.71) & 4.13(0.43) & 4.40(0.51) \\
    OpenAI-o3-mini \citep{O3} & 4.05(0.73) & 4.53(0.28) & \underline{4.20}(0.56) & \underline{4.43}(0.41) \\
    Gemini-2.0-flash \citep{team2023gemini} & 3.90(0.97) & 4.40(0.79) & 4.09(0.61) & 4.35(0.49)\\
    DeepSeek-R1\citep{deepseekai2025deepseekr1incentivizingreasoningcapability}  & \underline{4.45}(0.54) & \textbf{4.68} (0.49) & 3.70(1.07) & 4.05(0.81) \\
    \midrule
    Preacher (\textbf{Ours}) & \textbf{4.70}(0.35) & \underline{4.63}(0.34) & \textbf{4.38}(0.66) & \textbf{4.50}(0.31) \\
    \bottomrule
    \end{tabular}
    \vspace{-0.07in}
}
\end{table}
\subsection{Main Results}
\label{sec5_2}
\cref{tab:overall} compares Preacher with OpenAI o3-mini + state-of-the-art video generation models. Preacher outperforms existing methods in six out of ten metrics, notably in accuracy, professionalism, and alignment with the paper. Human evaluations further confirm Preacher’s superiority, as LMMs struggle to distinguish professional content in videos. Preacher’s use of domain-specific styles (e.g., mathematical visualizations, slide-based formats) may reduce scores in aesthetic quality and CLIP similarity, but this trade-off preserves scholarly integrity.

\cref{tab:overall2} evaluates Preacher’s key scene planning, where it leads in three out of four metrics. Chain-of-thought reasoning improves accuracy and professionalism but often results in overly complex scene plans, reducing compatibility with generative models.

\cref{fig:showcase} compares Preacher-generated video segments with those from OpenAI-o3-mini+Sora. OpenAI-o3-mini summarizes papers but lacks structured scene planning, leading to excessively complex textual descriptions that generative models struggle to process. General video generation models optimize for visual continuity and aesthetics but lack the domain-specific adaptability required for research content. In \cref{fig:showcase}(a), existing methods fail to adequately convey the concept of ``vector bundles'' and are unable to present the critical proof from the input paper. In \cref{fig:showcase}(b), although the prompt includes the crucial concept of the  ``lunar magnetic field,'' the excessive information in the prompt leads to incorrect generation, preventing the accurate representation of this important concept. 

By integrating progressive planning, multi-stage reflection mechanisms, and diverse video generation tools, Preacher ensures precise content representation, preventing the propagation of erroneous information in video abstracts.

\begin{figure}[h]
\centering
\includegraphics[width=0.90\linewidth]{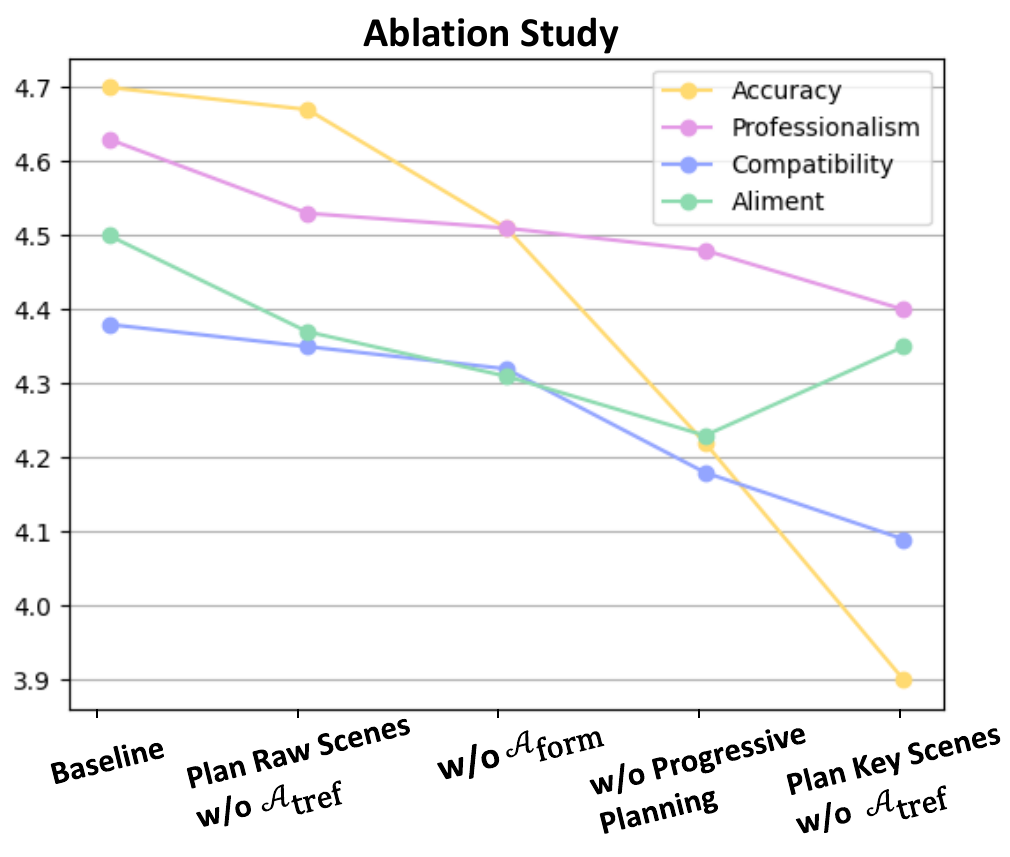}
\caption{Ablation Study on Preacher. ``w/o Progressive Planning'' refers to the planning of all components in key scenes at one time.}
\vspace{-0.17in}
\label{fig:abb_2}
\end{figure}
\subsection{More Analysis}
\paragraph{Ablation Study}
To assess the contribution of each mechanism in Preacher, we conducted comprehensive ablation studies, as shown in \cref{fig:abb_2}. Using Preacher as the baseline, we sequentially removed different mechanisms and evaluated the impact on key scene planning, following the same metrics outlined in \cref{sec5_2}.

Results in \cref{fig:abb_2} indicate that accurate key scene planning relies on the synergistic interaction of all mechanisms. Removing any component significantly reduces accuracy, while professionalism and compatibility exhibit lower sensitivity to such omissions. Notably, excluding the reflection mechanism during key scene planning improves alignment with the input paper. This is due to multi-round reflection causing scene drift, where iterative refinements lead to deviations from the original content. The progressive generation mechanism in Preacher mitigates this by iteratively incorporating the input paper and approved key scene components, ensuring that subsequent planning remains contextually anchored and prevents divergence.

\begin{figure}[h]
\centering
\includegraphics[width=0.90\linewidth]{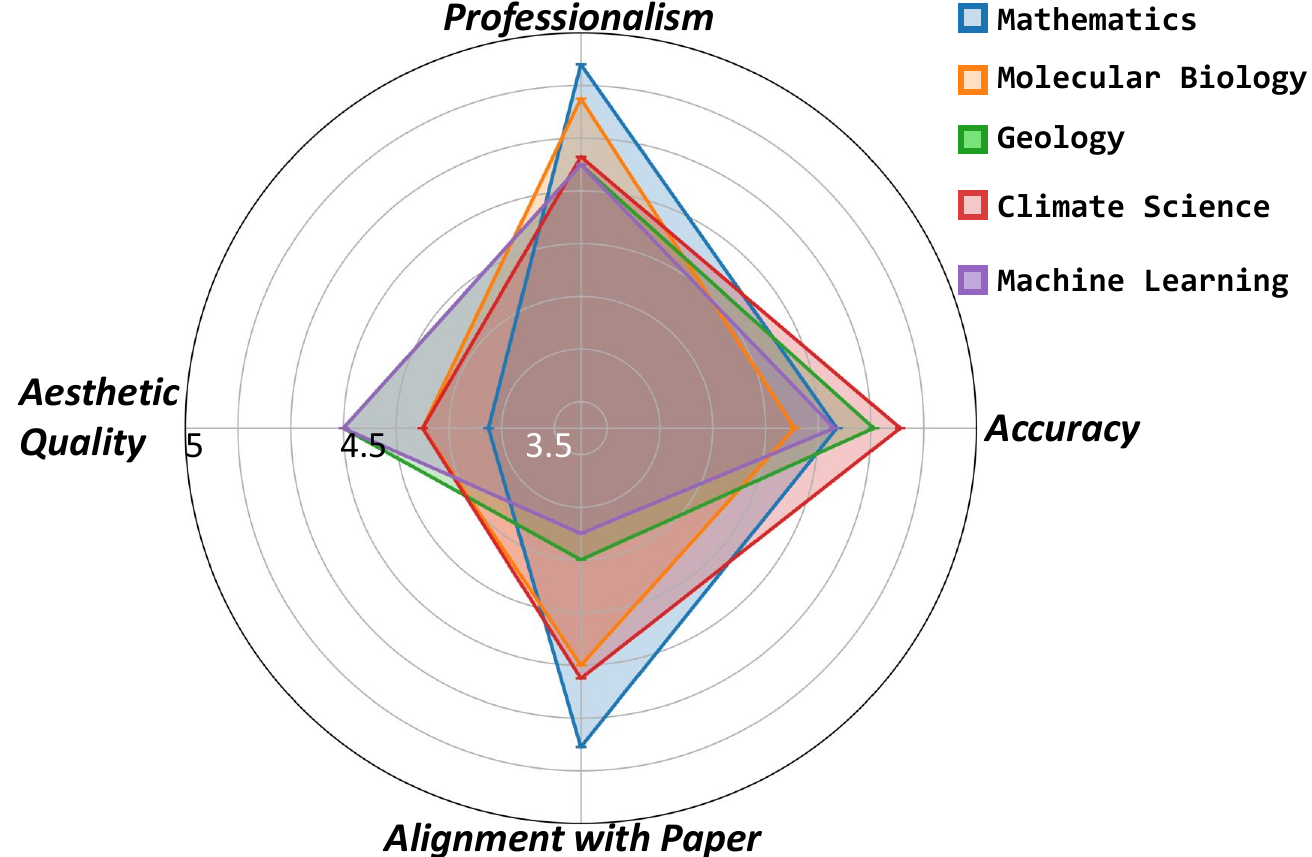}
\vspace{-0.07in}
\caption{Performance of Preacher with paper from different research fields.}
\vspace{-0.5em}
\label{fig:abb_1}
\vspace{-0.07in}
\end{figure}
\vspace{-1em}
\paragraph{Performance on Papers from Different Research Domains}
Preacher generates key scenes with diverse video styles, tailored to different research domains to ensure content alignment and effective knowledge dissemination. As shown in \cref{fig:abb_1}, these styles produce distinct visual effects, reflecting the unique requirements of various academic disciplines. While high evaluation scores are generally observed across styles, achieving simultaneous excellence in both professionalism and aesthetics remains challenging. This trade-off likely arises from Preacher’s prioritization of content accuracy, which inherently limits the complexity of visual composition and stylistic embellishments. Moreover, certain research fields, such as mathematics and molecular biology, require precise and schematic representations, further constraining the integration of elaborate visual effects. However, as text comprehension capabilities in video generation models continue to improve, allowing for a more balanced integration of scientific rigor and visual appeal.

\section{Conclusions and Limitations}
\paragraph{Conclusions} We introduce Preacher, the first paper-to-video agentic system. By leveraging a top-down and bottom-up agentic architecture, Preacher facilitates enhanced collaboration between agents. Through progressive chain-of-thought planning, Preacher systematically plans key scenes, generating high-quality video abstracts enriched with domain-specific expertise. Our evaluation across multiple research domains demonstrates Preacher’s effectiveness in representing and communicating domain-specific knowledge. In future work, we seek to broaden Preacher’s scope and applicability by integrating more video generation tools with diverse stylistic capabilities, ensuring adaptability to diverse disciplines and presentation formats.
\paragraph{Limitations}
As the first method to achieve paper-to-video generation, Preacher has several limitations.
First, its multi-agent collaboration necessitates over an hour for end-to-end processing, with token consumption for inter-agent communication.
Second, the absence of high-fidelity text-to-animation models restricts Preacher’s ability to generate animation-style content, limiting its visual versatility.
Lastly, when processing papers in fields like artificial intelligence, key scenes are confined to ``slides" and ``talking heads" due to the abstract nature of such papers, which primarily comprise methodological descriptions and experimental analyses rather than concrete visualizable concepts.

\vspace{-1.2em}
\section{Acknowledgments}
\vspace{-0.5em}
\label{sec_8}
This work is supported by the Damo Academy through Damo Academy Research Intern Program. This work is also supported by the National Natural Science Foundation of China (No.62172018, No.62102008) and Wuhan East Lake High-Tech Development Zone National Comprehensive Experimental Base for Governance of Intelligent Society.

\vspace{-0.3in}
{   \small
    \bibliographystyle{ieeenat_fullname}
    \bibliography{main}
}

\end{document}